\documentclass[10pt,twocolumn,letterpaper]{article}

\usepackage{cvpr}              

\usepackage{tikz}
\usepackage{pgfplots}
\pgfplotsset{compat=1.18} 
\usepackage{comment}
\usepackage{xcolor,colortbl}
\usepackage{tablefootnote}

\definecolor{cvprblue}{rgb}{0.21,0.49,0.74}
\usepackage[pagebackref,breaklinks,colorlinks,allcolors=cvprblue]{hyperref}

\usepackage{siunitx}
\usepackage[accsupp]{axessibility}  
\usepackage{subcaption}

\def\paperID{19} 
\def\confName{CVPR}
\def\confYear{2025}

\title{Two Views Are Better than One: Monocular 3D Pose Estimation with Multiview Consistency}

\author{Christian Keilstrup Ingwersen$^{1,2}$ \and Rasmus Tirsgaard$^1$ \and Rasmus Nylander$^2$  \and Janus Nørtoft Jensen$^1$ \and Anders Bjorholm Dahl$^1$ \and Morten Rieger Hannemose$^1$ \\
$^1$ Visual Computing, Technical University of Denmark \and $^2$ TrackMan A/S, Denmark \\
{\tt\small \{cin, rany\}@trackman.com, \{rhti, abda, jnje, mohan\}@dtu.dk}}

\begin{document}
\maketitle
\begin{abstract}
Deducing a 3D human pose from a single 2D image is inherently challenging because multiple 3D poses can correspond to the same 2D representation. 
3D data can resolve this pose ambiguity, but it is expensive to record and requires an intricate setup that is often restricted to controlled lab environments. 
We propose a method that improves the performance of deep learning-based monocular 3D human pose estimation models by using multiview data only during training, but not during inference. 
We introduce a novel loss function, \emph{consistency loss}, which operates on two synchronized views. This approach is simpler than previous models that require 3D ground truth or intrinsic and extrinsic camera parameters.
Our consistency loss penalizes differences in two pose \emph{sequences} after rigid alignment. 
We also demonstrate that our consistency loss substantially improves performance for fine-tuning without requiring 3D data. Furthermore, we show that using our consistency loss can yield state-of-the-art performance when training models from scratch in a semi-supervised manner.
Our findings provide a simple way to capture new data, \eg in a new domain. This data can be added using off-the-shelf cameras with no calibration requirements. We make all our code and data publicly available.
\end{abstract}
\begin{figure*}[t]
    \centering
    \includegraphics[width=0.76\textwidth]{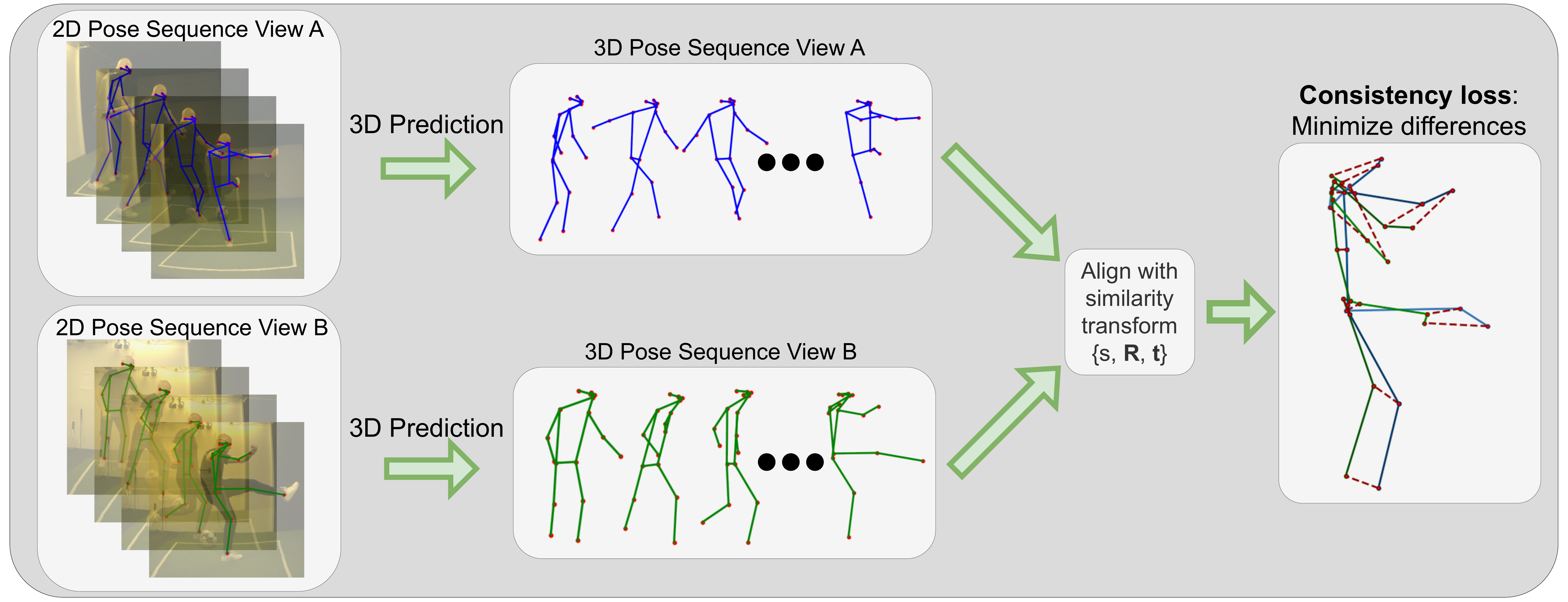}
    \caption{We improve monocular performance by applying our consistency loss during training to predicted 3D pose sequences from two different views. The consistency loss penalizes variations between the two predicted pose sequences of the same activity. Note that we only use multiple views during training. For every predicted 3D pose sequence obtained from View A and View B, we compute a similarity transform with Procrustes Analysis. This transformation aligns the predicted poses in Sequence A with Sequence B. The consistency loss is the average 3D distance between the two pose sequences post-alignment, shown as dashed red lines. Using Procrustes analysis for this transformation enables us to use cameras with unknown intrinsics and extrinsics.}
    \label{fig:consistency_overview}
\end{figure*}

\section{Introduction}
Inferring a 3D human pose from a single 2D image or 2D keypoints is an ill-posed problem, where the 3D pose is not uniquely defined. The consequence is that most existing models inaccurately predict the depth of the keypoints \cite{ingwersen2023evaluating}. It is, however, possible to learn a consistent mapping from a single 2D image to a 3D pose \cite{iqbal2020weakly,motionbert2022}. Previous methods train on 3D data \cite{motionbert2022} or on two or more views with a calibrated camera setup \cite{iqbal2020weakly}. Such data is expensive and restricts the settings where data can be recorded. Instead, we use a much simpler setup with two synchronized stationary or moving cameras. Here, the 3D pose in each frame is identical up to a transformation, and in the stationary case, this transformation is the same for all the frames recorded by the two cameras. We estimate this transformation using the Procrustes algorithm and optimize for consistent 3D pose predictions during training. At inference, we use a single 2D image, and our approach leads to state-of-the-art performance. \cref{fig:consistency_overview} sketches the concept of our consistency loss.

With this inherent ambiguity, models inferring 3D poses need a representation of how the body can move. It has become popular to rely on large foundation models~\cite{motionbert2022}, which have learned how the body moves in general based on supervised learning. Even so, foundation models do not always generalize to new movements in specific applications such as sports. This introduces the need for fine-tuning the foundation models to accommodate the less frequently seen movements specific to that domain~\cite{motionbert2022}, for which our model is particularly useful since collecting the data is inexpensive.

Methods for adapting 3D human pose models to different domains and movements have traditionally relied on the availability of new 3D data~\cite{Zhang_2022_CVPR, hmr, Zheng_2021_ICCV}. However, it can be costly and may not be feasible to set up systems for capturing 3D data in the new domain. Alternative methods have been explored to address these challenges. These methods have demonstrated that 3D pose models can be fine-tuned using 2D data, as suggested by previous work~\cite{Bogo:ECCV:2016}. This fine-tuning process ensures that the inferred 3D joint positions align with 2D keypoints in an image, which can be obtained accurately using readily available methods~\cite{2023arXivRTMPose, wang2021b}. Unfortunately, state-of-the-art models based solely on 2D supervision from a single view have too inaccurate depth predictions to be used in actual applications such as sports~\cite{ingwersen2023evaluating}. Thus, there is a need to advance 3D pose prediction from 2D images.

While we focus on increasing the performance of monocular models and not models utilizing multiple views, many datasets used for training monocular models have multiple views of the scene available~\cite{ingwersen2023sportspose, aspset, 3dhp, panoptic}. We use the multiple views in the SportsPose~\cite{ingwersen2023sportspose}, Human3.6M~\cite{h36m_pami}, and SkiPose~\citep{sporri2016reasearch} datasets to evaluate our approach. We use multiple views during training while only using a single view at inference. Additionally, we investigate how many views are necessary to obtain improvements in 3D predictions. Specifically, we use the SportsPose dataset~\cite{ingwersen2023sportspose} to investigate the performance using varying numbers of views. We have gained access to all seven views of this dataset, and the original authors have allowed us to release the additional data as part of this paper. The SportsPose dataset features complex and challenging sports scenarios, enabling us to test poses that are rare in other datasets. The new views are available on our website
\footnote{\url{http://christianingwersen.github.io/SportsPose}}.
Additionally, we test our consistency loss on the Human3.6M dataset trained semi-supervised, resulting in superior performance. By utilizing our consistency loss, we obtain state-of-the-art among semi-supervised methods.

Our contributions extend beyond introducing the view-consistency loss for domain-adaptive 3D pose estimation. We also present the first set of baseline results on the SportsPose dataset, demonstrating the effectiveness of our approach. We illustrate how our method enhances 3D pose estimation accuracy in dynamic and complex environments by showcasing a model fine-tuned on the SportsPose dataset. This research opens new possibilities for domain adaptation in 3D pose estimation, providing a practical and cost-effective solution to customize models for specific applications, and unlocks the possibility of increasing the state-of-the-art accuracy in monocular pose estimation by training on large amounts of two-view in-the-wild data.

\section{Related work}

\subsection{Monocular 3D human pose models}
\label{sec:monocular_3d_rel_works}
There are two primary approaches for monocular 3D human pose estimation. One solely predicts the 3D locations of the human skeleton~\cite{pavllo:videopose3d:2019, Zhang_2022_CVPR, shan2022p}, while the other includes estimating the body shape~\cite{xu2020ghum, Bogo:ECCV:2016, spin, hmr, metro}. Most models that include the body shape employ parametric body models such as SMPL~\cite{SMPL:2015}, which describes the body through shape parameters and pose parameters. Notably, our proposed loss remains applicable to both approaches, as 3D joint positions, as well as shape and pose parameters can be penalized based on the variation between multiple views.

Irrespective of whether the goal is to estimate pose alone or to include shape parameters, monocular 3D human pose estimation commonly adopts either a one-stage or a two-stage approach. In the one-stage approaches, the estimation is directly derived from an image or video input, while the two-stage approaches involve lifting estimated 2D poses to 3D space. State-of-the-art monocular models that employ the two-stage approach, lifting 2D poses to 3D, achieve mean-per-joint precision errors (MPJPE) as low as 17mm~\cite{motionbert2022} when lifting ground truth 2D poses on the Human3.6M dataset~\cite{h36m_pami}, and 37mm when lifting estimated 2D poses~\cite{motionbert2022}. However, when the same methods are evaluated on other datasets they have much higher MPJPE, it is clear that further work is still required~\cite{ingwersen2023evaluating}.

Models that adopt the alternative approach of inferring the 3D pose by estimating the parametric SMPL model directly from image input, have achieved  MPJPE scores of 60mm \cite{Shetty_2023_CVPR} on the in-the-wild 3DPW dataset~\cite{vonMarcard2018}.

\subsection{Multiview 3D human pose models and datasets}
Multiple synchronized and calibrated cameras have been extensively used to generate data to develop human pose estimation models~\cite{iskakov2019learnable, chun2023representation, Reddy_2021_CVPR}. Utilizing calibrated camera setups in such approaches has yielded impressive results, and has also been the basis for generating state-of-the-art 3D human pose datasets~\cite{ingwersen2023sportspose, aspset, 3dhp, panoptic}. These datasets have been essential for developing monocular pose models. However, the practical implementation of multi-camera setups involves a calibrated camera setup. Thus, most data has been collected in controlled laboratory environments, which does not reflect data variability in many scenarios.

Approaches that require limited or no 3D supervision have also been explored.
Some of these are unsupervised and train on single images by lifting a 2D pose to 3D followed by a random rotation and re-projection to 2D~\cite{drover2018can,chen2019unsupervised,wandt2022elepose,hardy2024links}. Others use multiple views of the same person~\cite{iqbal2020weakly, gholami2021tripose, mitra2020multiview, liu2023motion, kocabas2019self}, which is close to our approach. However, these approaches require camera calibration~\cite{liu2023motion}. Some methods only rely on intrinsic calibration and estimate relative camera poses by decomposing the essential matrix estimated from 2D poses predicted in multiple views. Then, the 3D poses are triangulated using the estimated relative poses and used as training data~\cite{kocabas2019self,gholami2021tripose}.
\citet{mitra2020multiview} add additional training data from multiview images and use metric learning to enforce that images of the same pose have similar embeddings. 
Similar to our approach, \citet{iqbal2020weakly} rigidly align 3D poses predicted from multiple views but opposed to our method they require known camera intrinsic and can only predict scale normalized poses.
During training, they penalize the model for differences in the predicted poses. 
Both \citet{iqbal2020weakly} and \citet{mitra2020multiview} apply multiview consistency on single image pairs and not sequences, 
thus risk reducing accuracy and potential use by ignoring temporal information.

\section{Multiview consistency loss}
In our approach, we propose a loss function that optimizes for consistency between a sequence of 3D poses predicted from multiple views. The consistency is measured as the distance between Procrustes aligned pose sequences. Therefore, we do not need to know the cameras' intrinsic or extrinsic calibration or other prior information. Instead, the consistency loss applies a similarity transformation and penalizes differences in the poses of two sequences of the same activity, see \Cref{fig:consistency_overview}. Avoiding camera calibration simplifies the training pipeline and gives an efficient alternative for handling data from multiple views.

Specifically, the loss is based on the difference between poses computed from two or more views after alignment with a similarity transformation, $\tau$. We compute the mean difference over every pair of two cameras, which results in the loss
\begin{align}
\mathcal{L}_\text{con}  = \sum_{s=1}^{S} \frac{1}{|V_s|}\sum\limits_{(a, b)\in V_s} \mathcal{L}_{\text{c}}\left(\hat{J}_a, \hat{J}_b\right).
\label{eq:consistency_loss_sum}
\end{align}
Here, $S$ is the total number of sequences, and $V_s$ is the set of possible pairs of views of the sequence $s$. Therefore, with $N$ different cameras available in a sequence, $|V_s| = \binom{N}{2}$. The consistency loss $\mathcal{L}_{\text{c}}$ is calculated between $\hat{J}_a$ and $\hat{J}_b$, which are the predicted 3D body joints for all frames of the sequence from view $a$ and view $b$, respectively. The term $\mathcal{L}_{\text{c}}$ is 
\begin{align}
\mathcal{L}_{\text{c}}(\hat{J}_a, \hat{J}_b) = \frac{1}{n} \sum_{i=1}^{n} \left\| \tau\left(\hat{J}_{a,i}; \hat{\theta}_{ab}\right) -  \hat{J}_{b, i} \right\|_2,
\label{eq:consistency_loss}
\end{align}
where $\hat{J}_{a,i}$ is element $i$, from the sequence of predicted 3D poses from view $a$, which has length $n$. Similarly, $\hat{J}_{b, i}$ is element $i$ from the sequence of predicted 3D poses from view $b$. $\tau$ is a similarity transform with parameters $\hat{\theta}_{ab}$ that are estimated such that $\tau$ transforms $\hat{J}_{a,i}$ to be as close as possible to $\hat{J}_{b,i}$ by scaling, rotating, and translating the 3D joints from $\hat{J}_{a,i}$.

To compute the scaling, rotation, and translation used to transform $\hat{J}_{a, i}$, we estimate the optimal parameters $\hat{\theta}_{ab}$, as in \Cref{eq:transform_param}. Here, it should be noted that contrary to how similarity transformations are traditionally computed in 3D human pose estimation to calculate the Procrustes aligned MPJPE, we only compute one transformation, $\hat{\theta}_{ab}$, for the entire sequence and not one per frame as in the PA-MPJPE metric~\cite{ingwersen2023evaluating}
\begin{equation}
\hat{\theta}_{ab} = \arg\min\limits_{\theta} \sum_{i=1}^{n} 
 \left|\left| \tau\left(\hat{J}_{a,i}; \theta\right) - \hat{J}_{b,i}\right|\right|_2^2 .
\label{eq:transform_param}
\end{equation}

The optimal solution to \Cref{eq:transform_param} is found using Procrustes analysis~\cite{procrustes}, such that we obtain the optimal scaling, rotation, and translation to transform $\hat{J}_{a,i}$ as follows
\begin{equation}
    \tau\left(\hat{J}_{a,i}; \hat{\theta}_{ab}\right) = s \hat{J}_{a,i} R + t.
    \label{eq:transform}
\end{equation}

By transforming $\hat{J}_{a}$, the idea is to directly estimate the similarity transformation that transforms from the camera coordinate system of camera $a$ to the coordinate system of camera $b$ instead of relying on knowing the camera extrinsics to perform the transformation.

\section{Experiments}
\label{sec:exp}
We conduct experiments with fine-tuning a pretrained pose estimator on a new dataset with ground truth 3D (\cref{sec:finetune_3d}) and without any 3D data (\cref{sec:finetune_2d}). In \cref{sec:h36m} we show the applicability of our loss when training from scratch in a weakly supervised setting. Finally, we investigate the number of views our consistency loss needs (\cref{sec:how_many_views,sec:more_views_or_more_data}).

\paragraph{Datasets}
To evaluate our method we have chosen several datasets that all contain multiple views and ground truth 3D poses.

\emph{Human3.6M}~\cite{h36m_pami} is the most widely used dataset in human pose estimation. It is recorded with four fixed cameras and a total of 3.6 million frames across seven subjects.

\emph{SkiPose} \cite{sporri2016reasearch,rhodin2018learning} is a domain-specific dataset of 12 sequences of alpine skiers recorded with six PTZ cameras with a total of 10,197 frames. This lets us test the performance of our loss in the challenging setting of non-fixed cameras. 

\emph{SportsPose} contains several movements common in sports that pretrained models struggle with~\cite{ingwersen2023sportspose}.
It is recorded with seven fixed cameras and has a total of 1.5 million frames, which lets us test the impact of the number of views on our loss.

\paragraph{SportsPose test protocol}
As \citet{ingwersen2023sportspose} do not provide a specified test protocol, 
we employ a test protocol inspired by Human3.6M~\cite{h36m_pami}, wherein subjects are distributed across sets to ensure that no subject appears in the same set.

We use subjects S04, S07, S09, S14, and S22 for validation. Subsequently, we employ subjects S06, S12, and S19 for testing. To focus on monocular performance, we opt to use only the currently available view, ``right'', during the testing and validation of the model. This decision streamlines the evaluation process, as we are interested in assessing the proficiency of the model when exposed to a single front-facing view. Examples of this view are in the first row of \Cref{fig:front_facing}.

\newcommand*{\vertcat}[1]{%
  \leavevmode
  \vbox{
    \offinterlineskip 
    \hbox{\includegraphics[width=\linewidth]{figures/activities/#1_fo}}%
    \hbox{\includegraphics[width=\linewidth]{figures/activities/#1_dl}}%
  }%
}

\begin{figure}[tb]
  \centering
  \begin{subfigure}{0.2\linewidth}
    \centering
    \vertcat{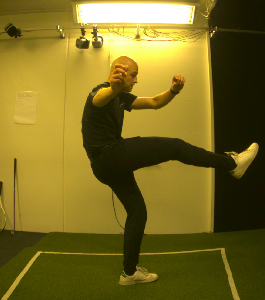}
    \caption*{Soccer}
  \end{subfigure}%
  \begin{subfigure}{0.2\linewidth}
    \centering
    \vertcat{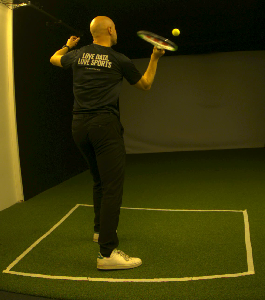}
    \caption*{Tennis}
  \end{subfigure}%
  \begin{subfigure}{0.2\linewidth}
    \centering
    \vertcat{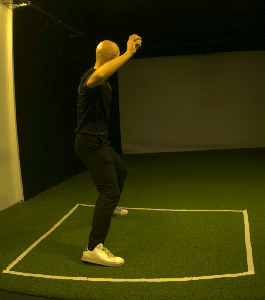}
    \caption*{Baseball}
  \end{subfigure}%
  \begin{subfigure}{0.2\linewidth}
    \centering
    \vertcat{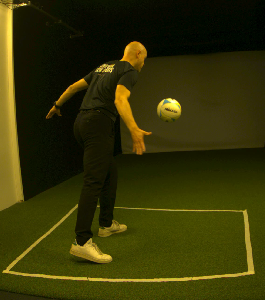}
    \caption*{Volley}
  \end{subfigure}%
  \begin{subfigure}{0.2\linewidth}
    \centering
    \vertcat{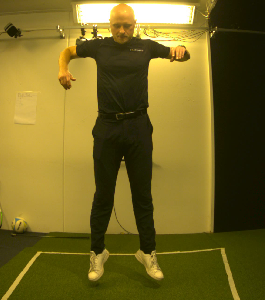}
    \caption*{Jump}
  \end{subfigure}%
  
  \caption{The five activities from SportsPose~\cite{ingwersen2023sportspose}. The top row displays the publicly available view ``right''. The bottom row features a view rotated  90 degrees relative to ``right'', which we refer to as ``View 1''.
  } %
  \label{fig:front_facing}
\end{figure}

\subsection{Implementation}
While our consistency loss is versatile and applicable to any monocular 3D human pose method, we choose to adapt the MotionBERT model by \citet{motionbert2022} due to its impressive performance on multiple datasets. For fine-tuning the SportsPose~\cite{ingwersen2023sportspose} dataset we used 2D poses predicted by RTMPose~\cite{2023arXivRTMPose}.

Details of the preprocessing of the detected 2D poses are in the supplementary material.

For the fine-tuning of MotionBERT~\cite{motionbert2022}, we employ the weights provided for the DSTformer with a depth of five and eight heads. The sequence length is 243, and the feature and embedding sizes are 512 as in the original paper. Adhering to the training protocol suggested by \citet{motionbert2022}, we fine-tune the models for 30 epochs, using a learning rate of 0.0002 and utilizing the Adam optimizer \cite{kingma2014adam}. 

\subsection{Fine-tuning with 3D data}
\label{sec:finetune_3d}

To compare to the situation when 3D data is available, we also experiment with fine-tuning with 3D data. We implement the proposed fine-tuning configuration from MotionBERT. This involves using a positional loss, $\mathcal{L}_{\text{pos}}$, directly on the 3D poses, coupled with losses on joint velocities, $\mathcal{L}_{\text{vel}}$, and scale only loss, $\mathcal{L}_{\text{scale}}$, as suggested by \citet{rhodin2018unsupervised}. This combination results in the combined loss for 3D data,
\begin{align}
    \begin{aligned}
        \mathcal{L}_{\text{3D}} = \lambda_{\text{pos}} \mathcal{L}_{\text{pos}} + \lambda_{\text{vel}} \mathcal{L}_{\text{vel}} + \lambda_{\text{scale}} \mathcal{L}_{\text{scale}},
    \end{aligned}
\label{eq:3d_combined_loss}
\end{align}
where $\lambda_{\text{pos}}$, $\lambda_{\text{vel}}$, and $\lambda_{\text{scale}}$ are weights for the respective losses. Our proposed consistency loss is added as a regularization term, $\lambda_{\text{con}} \mathcal{L}_{\text{con}}$, to the total loss, resulting in \Cref{eq:3d_combined_loss_total},
\begin{align}
    \begin{aligned}
        \mathcal{L}_{\text{3D}_\text{con}} = \lambda_{\text{pos}} \mathcal{L}_{\text{pos}} + \lambda_{\text{vel}} \mathcal{L}_{\text{vel}} + \lambda_{\text{scale}} \mathcal{L}_{\text{scale}} + \lambda_{\text{con}} \mathcal{L}_{\text{con}}.
    \end{aligned}
\label{eq:3d_combined_loss_total}
\end{align}

After an extensive parameter search, aligning with suggestions from \citet{motionbert2022}, we identify the optimal configuration for \Cref{eq:3d_combined_loss_total} as $\lambda_{\text{pos}} = 1$, $\lambda_{\text{vel}} = 20$, $\lambda_{\text{scale}} = 0.5$, and $\lambda_{\text{con}} = 0.2$. These parameters are employed to obtain the results presented in \Cref{tab:two-view-res}, utilizing two camera views from SportsPose~\cite{ingwersen2023sportspose}, one from the right side, as illustrated in the first row of \Cref{fig:front_facing}, and another 90 degrees to the view facing the back of the subject as in the second row of \Cref{fig:front_facing}. The second view behind the subject is based on the assumption that this view contains the most information when joints are occluding each other in the original ``right'' view from SportsPose~\cite{ingwersen2023sportspose}.

\sisetup{detect-weight=true}
\newcommand{\besttwoD}[1]{\best{\color{black}#1}}
\newcommand{\bestthreeD}[1]{\best{\color{gray}#1}}
\begin{table*}[btp]
    \centering
    \begin{minipage}{0.67\textwidth} 
\centering
\newcommand{\best}[1]{\textbf{#1}}
  \caption{Results on SportsPose~\cite{ingwersen2023sportspose}. Baseline is MotionBERT~\cite{motionbert2022}, which is then fine-tuned with either 2D ($\mathcal{L}_{\text{2D}}$) or 3D ($\mathcal{L}_{\text{3D}}$) supervision with and without our proposed multiview consistency loss $\mathcal{L}_{\text{con}}$. All results are in mm where lower is better. MPJPE is mean per joint precision error, and PA is Procrustes aligned MPJPE. All results use predicted 2D poses~\cite{2023arXivRTMPose}. \besttwoD{Bold} is the best performance with only 2D data and \bestthreeD{bold gray} is best performance with 3D ground truth. The two views are shown in \Cref{fig:front_facing}. The consistency loss improves performance for both 2D and 3D but substantially more when 3D supervision is not used.}
\setlength{\tabcolsep}{1.6pt}
\begin{tabular}{l *{10}{S[table-format=2.1]}*{2}{S[table-format=2.1]}}
\toprule
&  \multicolumn{2}{c}{Soccer}& \multicolumn{2}{c}{Tennis  } & \multicolumn{2}{c}{Baseball }\\
& \multicolumn{2}{c}{ kick} & \multicolumn{2}{c}{serve} & \multicolumn{2}{c}{pitch} & \multicolumn{2}{c}{Volley} & \multicolumn{2}{c}{Jumping} & \multicolumn{2}{c}{All} \\ 
& \scalebox{.6}{MPJPE} & \scalebox{.6}{PA}& \scalebox{.6}{MPJPE} & \scalebox{.6}{PA}& \scalebox{.6}{MPJPE} & \scalebox{.6}{PA}& \scalebox{.6}{MPJPE} & \scalebox{.6}{PA}& \scalebox{.6}{MPJPE} & \scalebox{.6}{PA}& \scalebox{.6}{MPJPE} & \scalebox{.6}{PA}\\
\midrule
\multicolumn{1}{@{}l}{\text{\bf Baseline}}  \\
 MotionBERT~\cite{motionbert2022} & 64.2 & 39.5 & 70.7 & 39.7 & 85.0 & 42.2 & 86.8 & 50.0 & 78.0 & 48.9 & 77.1 & 44.1       \\
 \citet{iqbal2020weakly}\tablefootnote{As they have not released their code, we re-implement an improved version of their method by incorporating MotionBERT as the backbone with sequence length 1.} &  42.8 & 28.5 & 39.9 & 26.2 & 47.0 & 30.1 & 40.3 & 27.9 & 44.9 & 30.1 & 42.9 & 28.5 \\
\midrule
\multicolumn{5}{@{}l}{\bf \text{Fine-tuning with 3D data (2 views)}} \\

$\mathcal{L}_{\text{3D}}$ \labelcref{eq:3d_combined_loss}  & 26.7 & \bestthreeD{20.2} & 27.3 & 20.1 & 30.1 & 22.5 & 31.3 & 24.2 & 27.9 & 21.4 & 28.7 & 21.7 \\
$\mathcal{L}_{\text{3D}_\text{con}}$\labelcref{eq:3d_combined_loss_total}, Ours  & \bestthreeD{26.1} & 20.5 & \bestthreeD{25.4} & \bestthreeD{18.8} & \bestthreeD{29.4} & \bestthreeD{22.4} & \bestthreeD{30.8} & \bestthreeD{23.8} & \bestthreeD{27.9} & \bestthreeD{20.9} & \bestthreeD{28.0} & \bestthreeD{21.3} \\
\midrule

\multicolumn{5}{@{}l}{\text{\bf Only 2D fine-tuning (2 views)}}  \\
 $\mathcal{L}_{\text{2D}}\labelcref{eq:2d_combined_loss}$ & 59.0 & 44.1 & 59.1 & 42.0 & 73.8 & 45.1 & 64.7 & 47.8 & 65.0 & 45.6 & 64.4 & 45.0       \\
$\mathcal{L}_{\text{2D}_\text{con}^\text{frame}}$, Ours & 
\besttwoD{33.9} & {21.9}       & \besttwoD{31.0} & \besttwoD{20.1}       & \besttwoD{36.1} & \besttwoD{23.0}           & {36.5} & {23.9}   & \besttwoD{34.1} & {23.5}       & \besttwoD{34.4} & \besttwoD{22.5}\\
$\mathcal{L}_{\text{2D}_\text{con}}$\labelcref{eq:2d_combined_loss_total}, Ours  & 
 {35.4} & \besttwoD{20.9}       & {36.2} & {22.7}       & {40.9} & {26.1}           & \besttwoD{33.5} & \besttwoD{22.6}   & {35.1} & \besttwoD{21.5}       & {36.2} & {22.8}      \\
\bottomrule
\end{tabular}
\label{tab:two-view-res}
\end{minipage}
    \hfill 
    \begin{minipage}{0.32\textwidth} 
    \centering
    \caption{Results on SkiPose~\cite{sporri2016reasearch} {\color{gray}Methods in gray} use all six views during inference, the rest are monocular. "F" indicates fine-tuning a pretrained model, "S" is self supervised using only 2D on SkiPose. Best performance without ground truth 2D annotations is marked in \textbf{bold}. }
        \begin{tabular}{@{}l@{}S[table-format=3.1]@{\hspace{3pt}}S[table-format=3.1]@{\hspace{3pt}}c@{}}
    \toprule
        Method & \text{\scalebox{.6}{MPJPE}} &  \text{\scalebox{.6}{PMPJPE}} & Uses \\ \midrule
        \color{gray}
        \color{gray}Metapose\cite{usman2022metapose} & \color{gray}\text{-} & \color{gray}42 & \color{gray}S \\ 
        \color{gray}\citet{zhou2023efficient} & \color{gray}42.2 & \color{gray}29.4 & \color{gray}F+3D\\
        MotionBERT \cite{motionbert2022} & 259 & 132 & \\ 
        CanonPose: \cite{wandt2021canonpose} & 128.1 & 89.6 & S \\ 
        MHCanonNet \cite{kim2024mhcanonnet}  & 122 & 50.7 & S \\ 
        \citet{yang2024geometry} & \text{-} & 68.4 & S \\ 
        \citet{kim2022cross} & 115.2 & 78.8 & S\\ 
        \citet{pavllo20193d}\textsuperscript{\ref{foot:reported_by}}  & 106 & 88.1 & F+S\\ 
        PoseAug \cite{gong2021poseaug}\tablefootnote{Value for \cite{gong2021poseaug} and \cite{pavllo20193d} reported by \cite{gholami2022adaptpose}.\label{foot:reported_by}} & 105.4 & 83.5 & F+S \\ 
        AdaptPose\cite{gholami2022adaptpose} & 99.4 & 83.0  & F+S\\ 
        $\mathcal{L}_{\text{2D}_\text{con}^\text{frame}}$, Ours & \bf62.0 & \bf40.5 & F+S \\ 
        $\mathcal{L}_{\text{2D}_\text{con}^\text{frame}}$, Ours, $\text{2D}^\text{GT}$ & 49.1 & 32.3 & F+S \\ 
        \bottomrule
    \end{tabular}
    \label{tab:skipose}
    \end{minipage}
\end{table*}

The results in \Cref{tab:two-view-res} show the impact of the consistency loss on model performance. Even when ground truth 3D data is available, the consistency loss yields marginal improvements, with a $0.8$mm decrease in MPJPE and $0.2$mm in PA-MPJPE. This slight enhancement suggests that our regularization term can be seamlessly integrated even when 3D data are available without compromising performance. It is crucial to emphasize that our consistency loss is intentionally crafted for scenarios lacking 3D data. But this underscores its utility as a valuable regularization technique for monocular pose estimation, acknowledging that the use of 3D data remains superior to achieve a well-performing model.

\subsection{Fine-tuning without 3D data}
\label{sec:finetune_2d}
Our consistency loss is especially useful when fine-tuning a model on a new dataset where multiple views are available but 3D poses are not.

The common way of utilizing additional data without 3D poses during training is to penalize deviations between the 2D poses and the reprojection of the predicted 3D poses~\cite{kanazawa2019learning}. When this is done with ground truth 2D it can work, but it can result in poor results due to the 2D-3D ambiguity~\cite{ingwersen2023evaluating}. Obtaining ground-truth 2D poses further requires manual annotations. Instead, predicted 2D poses are commonly used. We denote this 2D reprojection loss as
\begin{align}
    \begin{aligned}
        \mathcal{L}_{\text{2D}} = \lambda_{\text{2D}_{\text{reproj}}} \mathcal{L}_{\text{2D}_{\text{reproj}}}.
    \end{aligned}
\label{eq:2d_combined_loss}
\end{align}

We also add our consistency loss as regularization 
\begin{align}
    \begin{aligned}
        \mathcal{L}_{\text{2D}_\text{con}} = \lambda_{\text{2D}_{\text{reproj}}} \mathcal{L}_{\text{2D}_{\text{reproj}}} + \lambda_{\text{con}} \mathcal{L}_{\text{con}}.
    \end{aligned}
\label{eq:2d_combined_loss_total}
\end{align}
Through experimentation, we found that we achieve the best performance, with $\lambda_{2D_{\text{reproj}}} = 1$ and $\lambda_{\text{con}} = 0.3$, and use these values unless otherwise specified. 

Using the same two views as described in \Cref{sec:finetune_3d} we fine-tune on the SportsPose dataset with the loss from \Cref{eq:2d_combined_loss} or \Cref{eq:2d_combined_loss_total} and present the results in \Cref{tab:two-view-res}. Here we also present results with a variant of our consistency loss $\mathcal{L}_{\text{2D}_\text{con}^\text{frame}}$, where the optimal similarity transform is computed per frame, to allow for moving cameras. In \cref{tab:skipose} we present results on fine-tuning on the SkiPose dataset only with $\mathcal{L}_{\text{2D}_\text{con}^\text{frame}}$, as the cameras are not fixed.

\begin{figure}[btp]
  \centering
    \begin{tabular}{@{}c@{}c@{}c@{}c@{}}
        RGB & 45.3 mm & 28.0 mm & 20.5 mm \\
        \includegraphics[height=2.2cm]{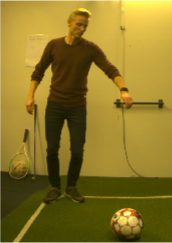} & 
        \includegraphics[height=2.2cm]{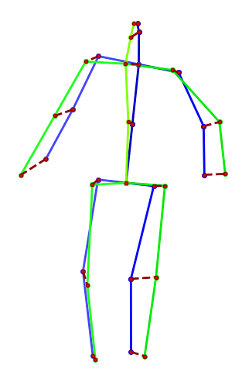} & 
        \includegraphics[height=2.2cm]{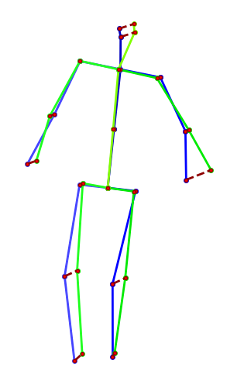} & 
        \includegraphics[height=2.2cm]{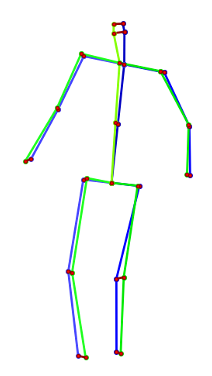} \\
        MPJPE: & 101.1 mm & 130.5 mm & 30.7 mm \\
        \includegraphics[height=2.2cm]{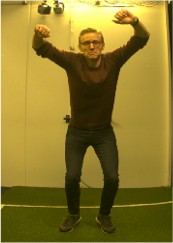} & 
        \includegraphics[height=2.2cm]{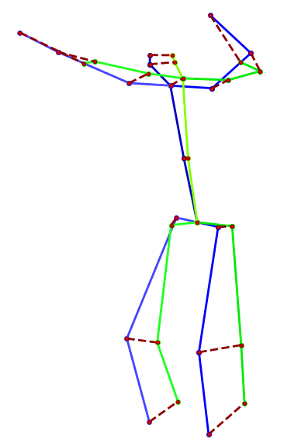} & 
        \includegraphics[height=2.2cm]{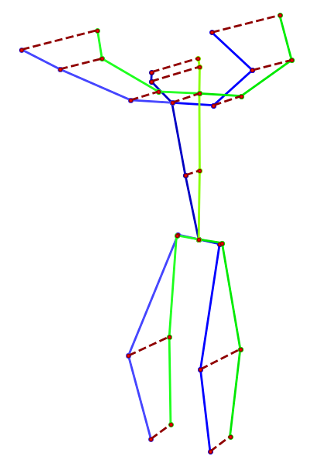} & 
        \includegraphics[height=2.2cm]{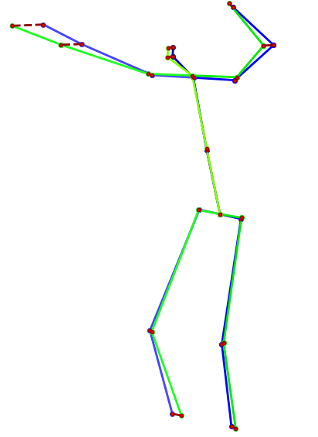} \\
        RGB & MotionBERT~\cite{motionbert2022} & $\mathcal{L}_{\text{2D}}$ & $\mathcal{L}_{\text{2D}_\text{con}}$, ours
    \end{tabular}
  \caption{Visual comparison of predictions in green and the ground truth pose in blue. The magnitude of errors, measured in millimeters and indicated at the top, highlights the superiority of our consistency loss $\mathcal{L}_{\text{2D}_\text{con}}$ in achieving more accurate results. The notable improvement is especially evident in the bottom row, where the method employing our consistency loss successfully captures the complex movement.}
  \label{fig:visual_errors}
\end{figure}

\Cref{tab:two-view-res,tab:skipose} highlights the impact of our consistency loss, particularly when 3D supervision is unavailable. The addition of the consistency loss on SportsPose leads to a substantial improvement in MPJPE, demonstrating a reduction of $39.2$mm compared to relying solely on the reprojection loss. \Cref{fig:visual_errors} shows predictions from models with and without the consistency loss and we see the same substantial increase in accuracy when using the consistency loss. 

\begin{table}[btp]
  \centering
    \caption{MPJPE and PA-MPJPE for different combinations of two views. The view ``right'' is included in all combinations. All experiments have been conducted with the loss $\mathcal{L}_{\text{2D}_\text{con}}$ and $\lambda_{\text{con}}=1$. It is clear that the two-view combination matters with right + view 1 and right + view 2 achieve substantially lower errors.}
  \begin{tabular}{ccc}\toprule
    \multicolumn{1}{@{}l}{Right $+$ view $x$}&\scalebox{1}{MPJPE} & \scalebox{1}{PA-MPJPE}\\\midrule
View 1  & 21.8 & \bf22.4 \\
View 2  & \bf21.6 & 24.3 \\
View 3  & 27.3 & 31.8 \\
View 4  & 25.6 & 26.7 \\
View 5  & 31.9 & 35.6 \\
View 6  & 25.8 & 27.2 \\\bottomrule
\end{tabular}

  \label{fig:error_plot}
  \end{table}

The improvement for 2D data is this big because the consistency loss has improved the network's ability to resolve ambiguities during the process of lifting 2D to 3D from a single view. In addition, our method proves beneficial in situations where joints might be occluded in one of the views, enhancing the overall robustness of the model.

However, a closer examination of the PA-MPJPE in \Cref{tab:two-view-res} reveals an interesting observation. Fine-tuning the model solely on 2D body keypoints based on \cref{eq:2d_combined_loss} results in an increase in error. A likely cause is that the 3D points that reproject to the same 2D points are not unique. Consequently, the model may struggle to provide accurate depth estimates of joint locations, as highlighted by \citet{ingwersen2023evaluating}. This underscores the importance of the consistency loss in mitigating such challenges and emphasizes its role in refining the performance of the model without ground truth 3D data.

\subsection{Comparison with state-of-the-art on Human3.6M data}
\label{sec:h36m}
\Cref{tab:h36m_comparison} shows results from our method with consistency loss compared with other semi-supervised methods trained on the Human3.6M~\cite{h36m_pami} dataset. For the evaluation, we adopt the same protocol as \citet{iqbal2020weakly} using 3D supervision from S1 and use S5, S6, S7, S8 with our consistency loss, $\mathcal{L}_{\text{2D}_\text{con}}$,  from \cref{eq:2d_combined_loss_total}. For evaluation, we use the standard protocol testing on subjects S9 and S11~\cite{h36m_pami, iqbal2020weakly}.

\begin{table}[bt]
\centering
\caption{Comparison with reported state-of-the-art Semi-Supervised methods trained from scratch on the Human3.6M~\cite{h36m_pami} dataset using only 3D data from S1 and weak supervision on S5, S6, S7, S8 during training. Unreported values are marked with "-". All 2D poses are predicted.}
\begin{tabular}{@{\hspace{5pt}} l@{} S[table-format=3.1]@{\hspace{5pt}} S[table-format=3.1]@{}}
\toprule
\multicolumn{1}{@{}l}{\textbf{Methods}}                   & \textbf{MPJPE$\,\downarrow$} & \textbf{PA-MPJPE$\,\downarrow$} \\
\midrule
\citet{rhodin2018unsupervised}~(ECCV'18)  & 131.7 & 98.2     \\
\citet{pavlakos2019texturepose}~(ICCV'19) & 110.7 & 74.5     \\
\citet{li2019boosting}~(ICCV'19)       & 88.8  & 66.5     \\
\citet{rhodin2018learning}~(CVPR'18)   &  \text{-}    & 65.1     \\
\citet{kocabas2019self}~(CVPR'19) & \text{-}  & 60.2     \\
\citet{iqbal2020weakly}~(CVPR'20)    & 62.8  & 51.4     \\
\citet{10044403}~(3DV'22) & 60.8 & 48.4 \\
\citet{Li_Pun_2023}~(AAAI'23) & 51.9 & \text{-} \\
Ours, $\mathcal{L}_{\text{2D}_\text{con}^\text{frame}}$ & 52.1 & 41.0 \\
Ours, $\mathcal{L}_{\text{2D}_\text{con}}$ \labelcref{eq:2d_combined_loss_total} & \bf50.5     & \bf40.4        \\
\bottomrule
\end{tabular}

\label{tab:h36m_comparison}
\end{table}

\Cref{tab:h36m_comparison} demonstrates that our novel consistency loss significantly improves performance in scenarios where 3D data is limited and with unlabeled multiview data. A key innovation of our approach, compared to \citet{iqbal2020weakly}, lies in incorporating the temporal aspect by computing a similarity transform per sequence instead of per frame, and not requiring known camera intrinsics. This advancement establishes a new state-of-the-art in semi-supervised performance on the Human3.6M dataset~\cite{h36m_pami}. Recent work \cite{10656673,gholami2022adaptpose} obtain very similar performance under the same weakly supervised protocol, but they utilize ground truth 2D poses, making the results not directly comparable.

\subsection{How many views do we need?}
\label{sec:how_many_views}
Examining the experiments carried out in \Cref{sec:finetune_3d,sec:finetune_2d}, a logical inquiry arises regarding the scalability of the results when more than two views are incorporated into the experiments. To investigate the correlation between the number of views and performance, we have calculated the results for scenarios where one to seven views are available, encompassing the total number of views in the SportsPose dataset~\cite{ingwersen2023sportspose}.

\textbf{Without available 3D data.}
In the absence of ground-truth 3D data, the influence of including multiple views on accuracy is evident as shown in \Cref{sec:finetune_2d} and the results in \Cref{tab:two-view-res}. To compute the results that involve more than two views without access to 3D data, we utilize the loss function from \Cref{eq:2d_combined_loss_total} with a consistent configuration, specifically setting $\lambda_{\text{2D}_\text{reproj}} = 1$ and $\lambda_{\text{con}} = 1$ for all experiments. It is essential to note that this configuration is not fine-tuned for a specific number of views, which may result in variations compared to the results presented in \Cref{tab:two-view-res}. The outcome of this ablation study is detailed in \Cref{fig:num_views_2d}.

Examining the results for 2D supervision in \Cref{fig:num_views_2d} reveals a substantial increase in accuracy as we progress from one to two views. However, the accuracy curve for both MPJPE and PA-MPJPE appears to plateau beyond two views, with marginal gains observed when incorporating more than two views. 

This observed plateau could be attributed to diminishing returns in information gain beyond the second view. While additional views contribute valuable perspectives, they may not necessarily introduce new information that significantly refines the precision of the predicted joints. Interestingly, this property of the loss underscores its utility, particularly in scenarios where capturing new data becomes significantly more manageable requiring only two views of the activity from an uncalibrated camera setup.

\textbf{With available 3D data.}
Even when 3D data is available, incorporating our consistency loss with two views results in a modest performance gain in MPJPE or PA-MPJPE, as illustrated in \Cref{sec:finetune_3d} and detailed in \Cref{tab:two-view-res}. This raises the question of whether this incremental gain will persist with an increasing number of views or reach a plateau, similar to the findings with only 2D.
In these experiments, we employ the loss function from \Cref{eq:3d_combined_loss_total} with $\lambda_{\text{pos}} = 1$, $\lambda_{\text{vel}} = 20$, $\lambda_{\text{scale}} = 0.5$, and $\lambda_{\text{con}} = 1$. Notably, these values are not fine-tuned for any specific number of views and may thus differ from the results presented in \Cref{tab:two-view-res}. The outcomes of this experiment are illustrated in \Cref{fig:num_views_3d}. We observe a slight increase in performance when additional views are added, along with the inclusion of our consistency loss. However, the variation in performance is generally small, and the overarching conclusion remains unchanged: when 3D data is available, there is no need to adapt the consistency loss.

\subsection{More views or more data?}
\label{sec:more_views_or_more_data}
Examining \Cref{fig:num_views_2d,fig:num_views_3d}, one may question if both the marginal accuracy improvements for 3D supervision and the substantial gains with 2D supervision with our consistency loss are only due to the increased amount of training data. To explore this we have conducted the same experiments but without including $\mathcal{L}_{\text{con}}$ in the loss.

\Cref{fig:num_data_2d} shows the experiment analogous to 2D supervision illustrated in \Cref{fig:num_views_2d}, but without the consistency loss. It  reveals that neither MPJPE nor PA-MPJPE exhibit improvement with the addition of more training data when incrasing the number of views. The observed plateau after two views contradicts the substantial accuracy increases depicted in \Cref{fig:num_views_2d}, suggesting that these improvements are attributed to the introduction of our consistency loss.

However, examining the experiments adding data to the 3D supervision in \Cref{fig:num_data_3d}, we observe a trend similar to that depicted in \Cref{fig:num_views_3d} with the error decreasing marginally when we add more data to the training. This suggests that the marginal improvements in accuracy when employing our consistency loss with 3D supervision can be attributed to the increased volume of data rather than solely to the presence of the consistency loss. 
This finding supports the overarching conclusion that 3D data is superior, and supports that the main advantage of our consistency loss lies in enhancing accuracy in scenarios where obtaining 3D data is impractical.

\begin{figure*}[tb]
    \newcommand{\xxlabel}{Number of views}
    \begin{subfigure}[t]{.48\textwidth}%
    \centering

    \begin{tikzpicture}
        \begin{axis}[
            xlabel={\xxlabel{}},
            ylabel={Error (mm)},
            xlabel style={yshift=4pt},
            ylabel style={yshift=-4pt},
            legend pos=north east,
            grid=major,
            width=\columnwidth,
            height=0.4\columnwidth,
            tick label style={font=\footnotesize},
            label style={font=\footnotesize},
            legend style={font=\footnotesize},
            ymin=0
            ]
            \addplot[color=blue,mark=square, line width=1pt] coordinates {
                (1,62.5)
                (2, 35.0)
                (3, 36.3)
                (4, 31.6)
                (5, 34.8)
                (6, 36.6)
                (7, 35.5)
            };
            \addplot[color=red,mark=triangle, line width=1pt] coordinates {
                (1, 44.5)
                (2, 22.0)
                (3, 22.0)
                (4, 20.7)
                (5, 21.8)
                (6, 22.4)
                (7, 21.1)
            };
        \end{axis}
    \end{tikzpicture}
    \caption{Loss: $\mathcal{L}_{\text{2D}_\text{con}}$. With just two available views, the error decreases significantly, but adding more views only decreases the error marginally.}
    \label{fig:num_views_2d}
    \end{subfigure}%
    \hfill
    \begin{subfigure}[t]{.48\textwidth}%
    \centering
    \begin{tikzpicture}
        \begin{axis}[
            xlabel={\xxlabel{}},
            ylabel={Error (mm)},
            xlabel style={yshift=4pt},
            ylabel style={yshift=-4pt},
            legend pos=south east,
            grid=major,
            width=\columnwidth,
            height=0.4\columnwidth,
            tick label style={font=\footnotesize},
            label style={font=\footnotesize},
            legend style={font=\footnotesize},
            ymin=0,
            ymax=30,
            ]
            \addplot[color=blue,mark=square, line width=1pt] coordinates {
                (1, 26.6)
                (2, 26.7)
                (3, 25.0)
                (4, 25.4)
                (5, 24.4)
                (6, 24.6)
                (7, 25.2)
            };
            \addplot[color=red,mark=triangle, line width=1pt] coordinates {
                (1, 20.6)
                (2, 20.5)
                (3, 19.4)
                (4, 19.1)
                (5, 18.8)
                (6, 19.1)
                (7, 18.9)
            };
        \end{axis}
    \end{tikzpicture}%
    \caption{Loss: $\mathcal{L}_{\text{3D}_\text{con}}$. There is a slightly lower error for more views, but the increase is far from as significant as when 3D data is not available.}%
    \label{fig:num_views_3d}%
    \end{subfigure}%
    \vskip\baselineskip
    \begin{subfigure}[t]{.48\textwidth}
    \centering
    \begin{tikzpicture}
        \begin{axis}[
            xlabel={\xxlabel{}},
            ylabel={Error (mm)},
            xlabel style={yshift=4pt},
            ylabel style={yshift=-4pt},
            legend pos=south east,
            grid=major,
            ymax=72,
            ymin=0,
            width=\columnwidth,
            height=0.4\columnwidth,
            tick label style={font=\footnotesize},
            label style={font=\footnotesize},
            legend style={font=\footnotesize},
            ]
            \addplot[color=blue,mark=square, line width=1pt] coordinates {
                (1, 62.9)
                (2, 61.2)
                (3, 59.3)
                (4, 57.8)
                (5, 57.7)
                (6, 57.7)
                (7, 59.7)
            };
            \addplot[color=red,mark=triangle, line width=1pt] coordinates {
                (1, 44.9)
                (2, 43.2)
                (3, 42.9)
                (4, 41.7)
                (5, 41.8)
                (6, 41.6)
                (7, 42.9)
            };
        \end{axis}
    \end{tikzpicture}
    \caption{Loss: $\mathcal{L}_{\text{2D}}$. The aim is to discern whether the increase in accuracy observed in \Cref{fig:num_views_2d} is influenced by the consistency loss or the augmented data availability. The nearly flat lines for both errors indicate that the accuracy boost associated with two or more views primarily stems from incorporating the consistency loss.}
    \label{fig:num_data_2d}
    \end{subfigure}
    \hfill
\begin{subfigure}[t]{.48\textwidth}
    \centering
    \begin{tikzpicture}
        \begin{axis}[
            xlabel={\xxlabel{}},
            ylabel={Error (mm)},
            xlabel style={yshift=4pt},
            ylabel style={yshift=-4pt},
            legend pos=south east,
            grid=major,
            width=\columnwidth,
            height=0.4\columnwidth,
            tick label style={font=\footnotesize},
            label style={font=\footnotesize},
            legend style={font=\footnotesize},
            ymax=30,
            ymin=0,
            ]
            \addplot[color=blue,mark=square, line width=1pt] coordinates {
                (1, 26.6)
                (2, 25.9)
                (3, 25.6)
                (4, 24.6)
                (5, 24.1)
                (6, 25.2)
                (7, 25.3)
            };
            \addlegendentry{MPJPE}
            \addplot[color=red,mark=triangle, line width=1pt] coordinates {
                (1, 20.6)
                (2, 19.9)
                (3, 19.8)
                (4, 19.0)
                (5, 18.4)
                (6, 19.5)
                (7, 19.5)
            };
            \addlegendentry{PA-MPJPE}
        \end{axis}
    \end{tikzpicture}
    \caption{Loss: $\mathcal{L}_{\text{3D}}$. The purpose is to explore whether the marginal improvements in accuracy in \Cref{fig:num_views_3d} are attributable to the consistency loss or the increased availability of data. Notably, we observe a slight decreasing trend in error as the number of views is increased, even without the consistency loss.}
    \label{fig:num_data_3d}
\end{subfigure}
    \caption{Investigation of how MPJPE and PA-MPJPE are affected when varying the number of views available with different losses. Top: With consistency loss, $\lambda_{\text{con}}=1$, bottom without consistency loss. Left: 2D, right: 3D.}
\end{figure*}

\subsection{Which views to choose}
In the experiments of \Cref{fig:num_views_2d,fig:num_views_3d}, the selection of views followed a deterministic process. Specifically, the first view was consistently chosen as the ``right'' view from SportsPose~\cite{ingwersen2023sportspose}, and the second view was facing the back of the subject \ie the one positioned closest to a 90-degree angle relative to the initial view. For scenarios involving three or more views, the remaining views were selected arbitrarily but maintained the same order across all experiments.

As the ambiguities of a 3D pose can theoretically be solved when observing from any two different views, a question arises if there is a practical advantage of certain viewpoints. This was investigated in \Cref{fig:error_plot} where the model was trained using the ``right'' view in combination with all other available views, where View 1 corresponds to the perspective positioned 90 degrees relative to the view facing the back of the subject.

Analyzing the errors depicted in \Cref{fig:error_plot}, it is evident that the choice of the view for multiview supervision significantly influences the outcomes of using our consistency loss. This aligns with intuition, as certain views are more effective in resolving ambiguities and identifying occluded joints, while others may not contribute new information. The results indicate that an optimal configuration involves using views from two sides that are 90 degrees apart when only two cameras are available.

\section{Discussion and conclusion}
\textbf{Limitations.}
While our results underscore a notable improvement in accuracy achieved through the implementation of our consistency loss, it is important to use the method with care.
As \Cref{fig:error_plot} demonstrates, the effect of the consistency loss depends on which views are used during training, with the least favorable combination resulting in performance comparable to using a single camera view. However, accuracy increases substantially in four out of five combinations.

Furthermore, emphasizing the consistency loss too much, indicated by a large $\lambda_{\text{con}}$ value, can lead to a degenerate solution. Specifically, an optimal solution to \Cref{eq:consistency_loss} may predict the same position of all joints.

Our proposed consistency loss estimates a single transformation for a pose sequence for aligning the 3D poses from different views. While this requires the cameras to be stationary throughout the sequence, our method can be extended to movable cameras by finding a similarity transformation for each time step.

It is worth noting that incorporating the proposed consistency loss necessitates temporal synchronization of pose sequences from different views. Desynchronisation will appear as correlated noise in the 2D predictions that in the worst case can confuse the model, especially for fast movements. This requirement imposes constraints on the camera system used for data capture. Using two cameras capturing frames with logged timestamps, it is possible to manually identify the same point in time in both sequences, or to use the audio to time-synchronize the views after acquisition~\cite{liang2017synchronization}. However, if using Android smartphones as cameras, the frame-synchronization can be obtained using an app and WiFi hotspots as described by \citet{Akhmetyanov2021SubmillisecondVS}.

While we demonstrate our loss on the SportsPose dataset~\cite{ingwersen2023sportspose}, which contains ground truth 3D data as well as a full multi-camera calibration, we only use the 3D data for evaluation, relying purely on predicted 2D keypoints for training. Because of the similarity transformation, our view-consistency loss eliminates the need for knowing the camera intrinsic or extrinsics, and only needs synchronized cameras. Foregoing camera calibration unlocks new opportunities for scaling multiview data acquisition as the footage can be captured on smartphones outside of the lab in diverse settings. The problem synchronization can be done with existing apps over WiFi~\cite{Akhmetyanov2021SubmillisecondVS} or solved as a post-processing step using audio cues~\cite{liang2017synchronization}.

\noindent\textbf{Conclusion.}
We present a novel method to enhance monocular 3D human pose estimation performance. By incorporating our multiview consistency loss during training in scenarios where 3D data is unavailable, we achieve notable performance improvements when compared to relying solely on 2D reprojection loss or no fine-tuning without requiring knowledge of the camera's extrinsic or intrinsic parameters.

We demonstrate the efficacy of the proposed consistency loss by evaluating it on the Human3.6M~\cite{h36m_pami}, SkiPose \cite{sporri2016reasearch}, and SportsPose~\cite{ingwersen2023sportspose} datasets. Following the semi-supervised protocol for Human3.6M, we advance the weakly supervised state-of-the-art precision. 

A thorough analysis exploring various configurations involving the number of views and camera placement reveals that an effective enhancement is achieved with just two appropriately positioned views. We observe that positioning the cameras at a 90-degree angle yields consistently good performance compared to other combinations of views. This demonstrates that, through the use of our multiview consistency loss, it is feasible to capture new domain data for fine-tuning a 3D model with a simple setup needing only two appropriately positioned and time-synchronized cameras.

With this paper, we also release six new views of sports activities to the SportsPose~\cite{ingwersen2023sportspose} dataset. Together with the new data we propose a new test protocol for the dataset and provide a simple baseline relying on MotionBERT~\cite{motionbert2022} and our proposed consistency loss.

{
    \small
    \bibliographystyle{ieeenat_fullname}
    \bibliography{main}
}

\end{document}


\maketitle

\section{Introduction}
The supplementary material contains details on the datasets and how they were preprocessed. Furthermore, additional detailed results are provided focusing on training with 2D vs.\ 3D ground truth points. The code is available from: [URL to come] 

\section{2D predictions}
For the Human3.6M, SportsPose, and Ski-Pose datasets the predicted 2D poses were obtained with RTMPose-L~\cite{2023arXivRTMPose} at a resolution of $384\times 288$ trained on a collection of datasets called Body8 as detailed on their \href{https://github.com/open-mmlab/mmpose/tree/main/projects/rtmpose}{Github page}.

\section{Dataset details}
\subsection{SportsPose}
\paragraph{Preprocessing}
To use SportsPose~\cite{ingwersen2023sportspose} with MotionBERT~\cite{motionbert2022} we need to preprocess the data. We convert the definition of body keypoints from COCO~\cite{cocodataset} keypoints used in SportsPose to the Human3.6M~\cite{h36m_pami} body keypoint format. The keypoints are further transformed from meters to millimeters and, by using the extrinsic camera parameters, transformed from the world coordinate system to each camera. Then, following the approach of \citet{ci2020locally}, the camera coordinates are transformed to pixel coordinates and scaled to be within the range $[-1;1]$.

\subsection{SkiPose}
\paragraph{Preprocessing}
The markers recorded in SkiPose\cite{rhodin2018learning} are in H36M format. 
We convert the 2D detected joints with the same pipeline as SportsPose, except we skip cropping to the person, as the images are already cropped.
As our 2D joint detector is in COCO format, we convert from COCO to H36M using the same method as for SportsPose. One exception is joint 9 (head) placement is identical to 10 (top-head) in SkiPose, so we mirror this when converting the joints. 
SkiPose contains missing frames in some sequences, so we impute the 2D joint detections of missing frames as the last existing detection and set the confidence to 0. More model-specific approaches would improve our results, such as exploiting the positional encoding in MotionBert, but we want to keep the approach model-agnostic.
Finally, we transform the keypoints from meters to millimeters.

\paragraph{Training}
We used the same settings as when fine-tuning on SportsPose, except we trained for 250 epochs with a learning rate decay of 0.995 to accommodate the smaller dataset size. 

\section{Training with 2D vs. 3D}
This section investigates how much information is added to the model when training on data consisting of sequences viewed by two cameras with only predicted 2D poses compared to a sequence with ground truth 3D. For this, we again fine-tune the model on SportsPose~\cite{ingwersen2023sportspose} but vary the number of subjects present in the training set from 1 to 15. The two views we use are right + View 1. \Cref{fig:subjectstrainedon} shows how the performance improves the more subjects are added. The improvement in MPJPE is more noisy than for PA-MPJPE, but the overall trend that more data improves performance is clear.

\begin{figure}[h]
  \centering
  \newcommand{\heig}{0.7\columnwidth}
  \begin{subfigure}{.49\textwidth}
    \begin{tikzpicture}
      \begin{axis}[
          xlabel={Number of training subjects},
          ylabel={MPJPE},
          legend pos=north east,
          grid style=dashed,
          width=\columnwidth,
          height=\heig,
          tick label style={font=\footnotesize},
          label style={font=\footnotesize},
          legend style={font=\footnotesize},
        ]
        
        \addplot[color=blue, mark=square]
        coordinates {
          (1, 50.52) (2, 49.30) (3, 39.56) (4, 44.92) (5, 40.25) (6, 37.53) (7, 42.19) (8, 37.11) (9, 37.62) (10, 38.06) (11, 35.98) (12, 35.86) (13, 34.76) (14, 33.36) (15, 36.52)
        };
        \addlegendentry{$\mathcal{L}_{\text{2D}_\text{con}}$}
        
        \addplot[color=red, mark=o,]
        coordinates {
          (1, 46.88) (2, 39.69) (3, 35.03) (4, 33.24) (5, 31.77) (6, 31.76) (7, 31.17) (8, 28.76) (9, 27.27) (10, 27.42) (11, 27.49) (12, 29.35) (13, 28.07) (14, 27.77) (15, 26.41)
        };
        \addlegendentry{$\mathcal{L}_{\text{3D}_\text{con}}$}
      \end{axis}
    \end{tikzpicture}
    \caption{MPJPE}
  \end{subfigure}
  \hfill
  \begin{subfigure}{.49\textwidth}
    \begin{tikzpicture}
      \begin{axis}[
          xlabel={Number of training subjects},
          ylabel={PA-MPJPE},
          legend pos=north east,
          grid style=dashed,
          width=\columnwidth,
          height=\heig,
          tick label style={font=\footnotesize},
          label style={font=\footnotesize},
          legend style={font=\footnotesize},
        ]
        
        \addplot[color=blue, mark=square,]
        coordinates {
          (1, 34.92) (2, 32.36) (3, 28.49) (4, 27.12) (5, 26.20) (6, 25.24) (7, 25.86) (8, 23.99) (9, 24.26) (10, 23.42) (11, 23.62) (12, 23.16) (13, 22.50) (14, 23.47) (15, 22.66)
        };
        \addlegendentry{$\mathcal{L}_{\text{2D}_\text{con}}$}
        
        \addplot[color=red, mark=o,]
        coordinates {
          (1, 33.36) (2, 29.79) (3, 26.10) (4, 24.94) (5, 23.22) (6, 23.25) (7, 23.08) (8, 21.97) (9, 20.36) (10, 20.55) (11, 20.41) (12, 21.88) (13, 20.50) (14, 20.63) (15, 20.12)
        };
        \addlegendentry{$\mathcal{L}_{\text{3D}_\text{con}}$}
        
      \end{axis}
    \end{tikzpicture}
    \caption{PA-MPJPE}
  \end{subfigure}
  
  \caption{Performance on SportsPose as a function of number of subjects used for training.}
  \label{fig:subjectstrainedon}
\end{figure}

Now, let each line be described by a function of the number of subjects, $E_{2D}(\cdot)$ or $E_{3D}(\cdot)$. For a point on the $\mathcal{L}_{\text{2D}_\text{con}}$ line trained on $s_{2D}$ subjects with error $E_{2D}(s_{2D})$, we find which number of subjects is necessary to obtain the same performance using 3D supervision by moving left along the horizontal $E_{2D}(s_{2D})$ line and compute the first intersection with the $\mathcal{L}_{\text{3D}_\text{con}}$ line. The number of subjects in the intersection is denoted $s_{3D}(s_{2D})$, and since it is an intersection $E_{2D}(s_{2D}) = E_{3D}(s_{3D}(s_{2D}))$. Then we compute the fraction $\frac{s_{3D}(s_{2D})}{s_{2D}}$ and compute the average for all points with intersections.

The average fraction is 31\% for MPJPE and 55\% for PA-MPJPE, from which we can infer that we need about three times as much data to reach similar performance when we compare data observed with two cameras and only predicted 2D joints against sequences with ground truth 3D available. To obtain a similar PA-MPJPE we only need approximately twice as many sequences.

{
    \small
    \bibliographystyle{ieeenat_fullname}
    \bibliography{main}
}